\documentclass[sigconf]{acmart}
\usepackage{algorithm}
\usepackage{algpseudocode}
\usepackage{amsmath}

\usepackage{amssymb}
\usepackage{latexsym}
\usepackage{graphicx}

\DeclareMathOperator*{\argmax}{argmax}
\algnewcommand{\LeftComment}[1]{\Statex \(\triangleright\) #1}

\newcommand\BibTeX{B\textsc{ib}\TeX}

\graphicspath{ {./images/} }

\AtBeginDocument{%
  \providecommand\BibTeX{{%
    \normalfont B\kern-0.5em{\scshape i\kern-0.25em b}\kern-0.8em\TeX}}}

\copyrightyear{2021}
\acmYear{2021}
\setcopyright{usgovmixed}\acmConference[ICMI '21]{Proceedings of the 2021 International Conference on Multimodal Interaction}{October 18--22, 2021}{Montréal, QC, Canada}
\acmBooktitle{Proceedings of the 2021 International Conference on Multimodal Interaction (ICMI '21), October 18--22, 2021, Montréal, QC, Canada}
\acmPrice{15.00}
\acmDOI{10.1145/3462244.3479925}
\acmISBN{978-1-4503-8481-0/21/10}

\begin{document}

\title[Decision-Theoretic Question Generation]{Decision-Theoretic Question Generation for Situated Reference Resolution: An Empirical Study and Computational Model}

\author{Felix Gervits}
\affiliation{%
  \institution{DEVCOM Army Research Laboratory}
  \city{Burlington}
  \state{MA}
  \country{USA}
}
\email{felix.gervits.civ@army.mil}

\author{Gordon Briggs}
\affiliation{%
  \institution{U.S. Naval Research Laboratory}
  \city{Washington, DC}
  \country{USA}
  }
\email{gordon.briggs@nrl.navy.mil}

\author{Antonio Roque}
\affiliation{%
  \institution{Tufts University}
  \city{Medford}
  \state{MA}
  \country{USA}
}
\email{antonio.roque@tufts.edu}

\author{Genki A. Kadomatsu}
\authornote{The work was conducted as part of an internship at ARL.}
\affiliation{%
 \institution{Tufts University}
 \city{Medford}
 \state{MA}
 \country{USA}
}
\email{genki.kadomatsu@tufts.edu}

\author{Dean Thurston}
\affiliation{%
 \institution{Tufts University}
 \city{Medford}
 \state{MA}
 \country{USA}
}
\email{dean.thurston@tufts.edu}

\author{Matthias Scheutz}
\affiliation{%
 \institution{Tufts University}
 \city{Medford}
 \state{MA}
 \country{USA}
}
\email{matthias.scheutz@tufts.edu}

\author{Matthew Marge}
\affiliation{%
  \institution{DEVCOM Army Research Laboratory}
  \city{Adelphi}
  \state{MD}
  \country{USA}
}
\email{matthew.r.marge.civ@army.mil}

\renewcommand{\shortauthors}{Gervits et al.}

\begin{abstract}
Dialogue agents that interact with humans in situated environments need to manage referential ambiguity across multiple modalities and ask for help as needed. However, it is not clear what kinds of questions such agents should ask nor how the answers to such questions can be used to resolve ambiguity. To address this, we analyzed dialogue data from an interactive study in which participants controlled a virtual robot tasked with organizing a set of tools while engaging in dialogue with a live, remote experimenter. We discovered a number of novel results, including the distribution of question types used to resolve ambiguity and the influence of dialogue-level factors on the reference resolution process. Based on these empirical findings we: (1) developed a computational model for clarification requests using a decision network with an entropy-based utility assignment method that operates across modalities, (2) evaluated the model, showing that it outperforms a slot-filling baseline in environments of varying ambiguity, and (3) interpreted the results to offer insight into the ways that agents can ask questions to facilitate situated reference resolution.  
\end{abstract}



\maketitle

\section{Introduction}
\label{sec:introduction}

Effective human-agent interaction is critical for many application domains ranging from customer service to space missions. 
In non-laboratory environments these agents frequently encounter novel concepts (and words, objects, actions, etc.) with which they are unfamiliar. This is unavoidable due to difficulties in collecting and labeling training data, but the problem is magnified when agents are placed in novel and unexplored environments. For example, a robot for space exploration may encounter objects and procedures that do not exist on Earth. Moreover, ambiguity often arises in multiple modalities (e.g., language and vision), and so agents will need mechanisms to resolve ambiguity across all of these modalities. 

One solution to this problem is to enable agents to engage in dialogue and ask questions of a human interlocutor to reduce ambiguity. This is particularly important for the task of reference resolution, in which an agent needs to identify a target referent that it may know little or nothing about. However, not all questions that agents ask are equally effective or efficient, and we want those that ask \textit{good} questions \cite{cakmak2012designing}. To understand what makes good questions, we 
analyzed data from a corpus in which people controlled a robot to carry out a collaborative tool organization task while asking questions to a commander (an experimenter). The corpus task was designed to contain high amounts of referential ambiguity across multiple modalities with respect to entities in the environment, and thus to elicit as many questions as possible. This builds on the assumption that the kinds of questions people ask in this task are good models for dialogue agents in collaborative task domains. 

The contributions of this paper are twofold. First, we conducted several empirical analyses of the dialogue data from the corpus. This included categorizing the distribution of question types used in the task, and exploring the effect of various dialogue-level factors on question generation, including speaker initiative and instruction granularity. Despite being implicated in the clarification process, few if any studies have explored the role that these factors play, especially in the context of a human-robot interaction (HRI) task. Next, from the dialogue data in the corpus, we developed a computational model of situated reference resolution for dialogue agents using a decision network. Decision-theoretic approaches have shown promise in modeling some aspects of conversational grounding under multi-modal uncertainty \cite{horvitz1999computational,paek2000uncertainty}. However, to our knowledge they have not been employed for the task of generating questions for reference resolution. The advantage of such an approach is that it is effective without pre-training, intrinsically explainable, and robust to uncertainty across multiple modalities. Our model is also generalizable in that it dynamically generates a decision network from perceptual input and parameterizes the network based on the entities and properties in the domain in real time. This enables agents to operate in unseen environments for which they have limited or no knowledge - a property that is rare among current models. 

In the next section, we survey work that studies how people ask and respond to questions, and what approaches exist for modeling question generation for situated reference resolution. In Section~\ref{sec:study}, we describe our empirical study to investigate how robots should ask questions to support disambiguation of novel entities, and present the results. In Section~\ref{sec:model} we introduce our decision-theoretic approach informed by the study data, and we evaluate it in Section~\ref{sec:evaluation}. 
Finally, contributions and future work are discussed in Section~\ref{sec:discussion}.

\section{Background} 
\label{sec:background}

In tasks where one dialogue partner is trying to learn 
from another, and more broadly in any dialogue task where grounding occurs, 
clarification is an essential process. 
Clarification primarily emerges in dialogue where common ground contributions are in the process of being 
established \cite{clark1996using} 
or when there is a misunderstanding or non-understanding \cite{hirst1994repairing}. Recovery strategies from these grounding failures have been theorized and studied \cite{paek2003toward,skantze2005exploring}. As a special case of grounding, 
reference resolution entails a collaboration where a 
listener tries to determine the \textit{referent} that a speaker intends \cite{clark1986referring}. \textbf{Our work investigates the decision-making that an agent would make toward optimal clarifications when trying to resolve
referents with a human in a collaborative way.}

\subsection{Question Selection and Clarification in Situated Human-Agent Dialogue}
Humans often use a variety of question forms to facilitate mutual understanding in task settings, so it is natural for situated agents to ask questions of their human interlocutors. One past focus has involved robots. Indeed, robot-generated clarification requests are increasingly being used to facilitate data collection for active learning algorithms \cite{bullard2018towards,thomason2020jointly}. \cite{chernova2014robot} provided guidelines for robot learning, including that the robot must be able to ask questions that reflect its mental model, such as missing information, task uncertainty, etc. This raises the question of what robot question-asking strategies are most effective for facilitating successful learning. Addressing this question is the focus of the analysis in Section~\ref{sec:study}. 

The impact of dialogue factors on question generation has not been adequately explored to our knowledge. For example, it is known that factors such as \textit{speaker initiative} are related to role \cite{hulstijn2003roles} and interactivity \cite{core2003role}, which has implications for how questions are asked. Initiative has been explored in dialogue and discourse modeling \cite{walker-whittaker-1990-mixed}, and in its effects on dialogue management in various domains \cite{chu1999form,graesser2005autotutor}, as well as in HRI studies \cite{baraglia2016initiative,munzer2017impact}. Nevertheless, we are unaware of any empirical studies that manipulated initiative to investigate question generation. Another factor that may influence clarifications is \textit{instruction granularity}. Instructions can be produced at different levels of granularity, ranging from discrete low-level commands to high-level plans. Though \cite{marge2020lets} found that people are more likely to use low-level commands when instructing a robot versus a human, robots still need to be equipped with the capability to respond to high-level intentions \cite{williams2015going}. Few if any studies have examined the role that speaker initiative and instruction granularity play in the process of clarification generation, especially in novel environments characterized by multi-modal ambiguity.

\subsection{Situated Reference Resolution}

The task of asking good questions relates to the body of work
on situated reference resolution. Early work established requirements
and information sources for multi-modal reference
resolution~\cite{devault-etal-2005-information, fransen2007using, gross2017reliability, kruijff2007situated, traum2012incremental}. Graph-based
approaches that represent referents and their semantic properties
against an environment have shown promise but are generally restricted
to controlled environments such as tabletops or game 
boards \cite{kennington2017simple,liu2013modeling,zender2009situated}. 

Probabilistic models are another popular approach \cite{funakoshi2012unified,gorniak2005probabilistic,williams2016situated}. While effective at identifying referents from a set of entities, they may lead to dispreferred questions, compared to approaches that leverage features or knowledge about the environment \cite{williams2019dempster}. The same drawback holds for POMDP-based approaches \cite{lison2013model,misu2010modeling,williams2007partially}. While they are generally quite effective for dialogue state tracking, they are less effective at generating clarification questions for collaborative reference resolution, especially in novel environments. This is because POMDPs typically require a predefined state-action space, but the relevant properties of a given environment cannot always be known in advance. As a result, these approaches tend to struggle with open-world interaction. Neural approaches have recently been developed for situated, multi-modal dialogue \cite{huang2021joint,kalpakchi2019spacerefnet,kleingarn2019speaker,moon2020situated,schlangen2015resolving}. While promising, these approaches typically require large, labeled training sets and are generally limited to the specific domains in which they were trained.

The present work most closely relates to the body of work on clarification for situated, collaborative reference resolution. Some work has explored the use of decision-theoretic approaches for this purpose, mainly for selecting questions that maximize expected utility \cite{bullard2018towards,komatani2017question,prodanov2005decision}. Information-theoretic approaches have also been developed which rely on selecting queries that reduce entropy over a set of groundings in an instruction \cite{deits2013clarifying,hemachandra2014information}. However, we are not aware of work that fuses the two approaches. Our model uses decision networks supplemented with an entropy-reduction method over object properties to enable an agent to initiate referential clarifications when there are multiple possible strategies. This promises to be (1) more generalizable and scalable than the algorithms described above because of its focus on decision-making over detailed object representations, (2) flexible enough to be used in open-world scenarios, and (3) natural, since the model is explicitly grounded in empirical studies of human-human dialogue.

\section{HuRDL Corpus Analysis}
\label{sec:study}

The HuRDL (Human-Robot Dialogue Learning) Corpus \cite{gervits2021sigdial} was used as the basis for empirical investigation of question generation. The corpus task was designed to investigate how robots should ask questions to learn novel entities when facing multi-modal ambiguity. A question-type analysis was performed on the dialogue data to examine the frequency of various question types, including \textit{WH-questions} (\textit{WHQs}; who, what, where, when, why, how) and \textit{Yes/No-Questions} (\textit{YNQ}). These results served as the empirical basis for the computational model described in Section~\ref{sec:model}. Additional analyses examined the effect of dialogue-level factors on question generation (see Section~\ref{sec:results}). 

\subsection{Task Overview} 
\begin{figure}[t]
	\centering
	\includegraphics[width=\linewidth]{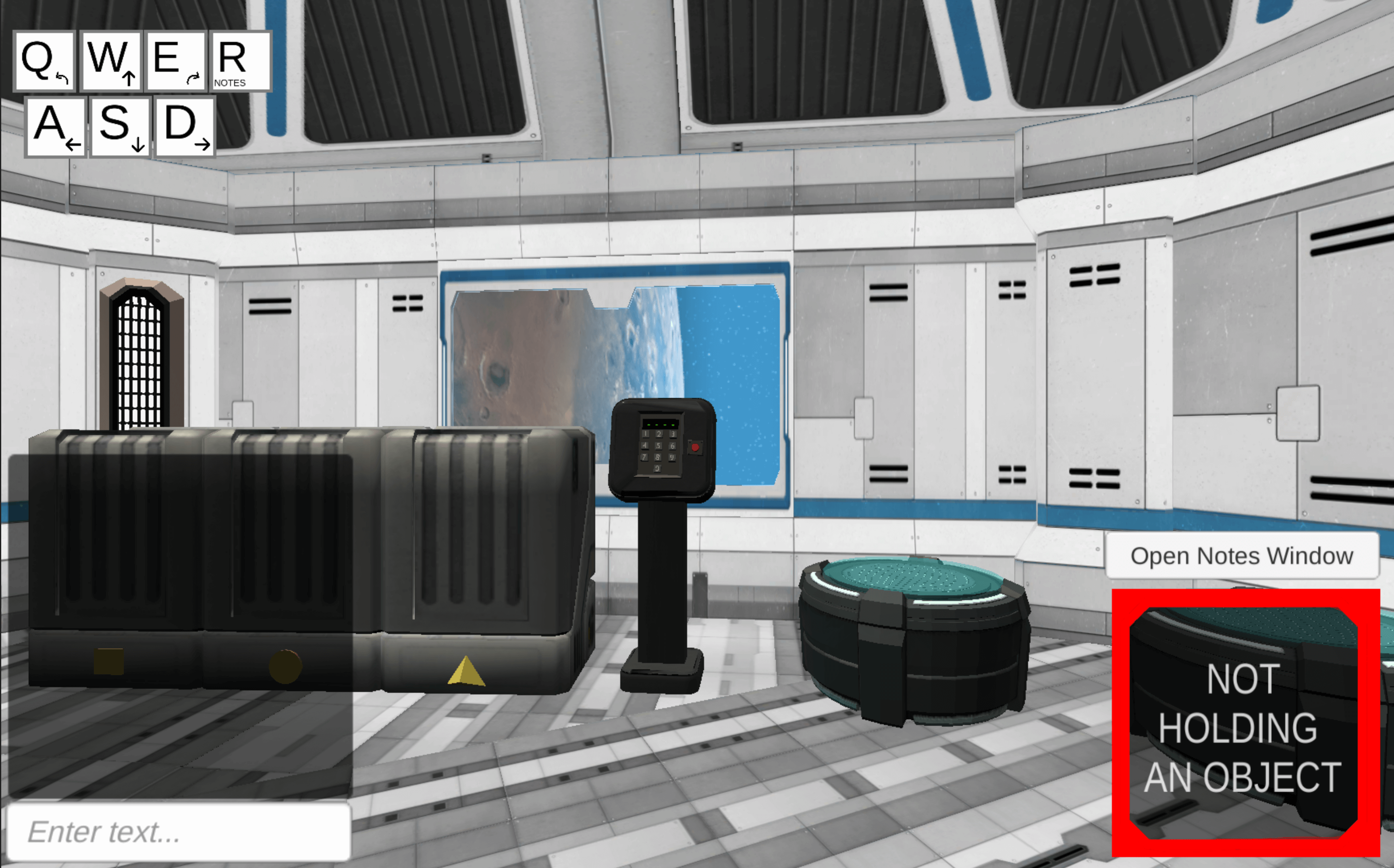}
	\caption{Participant view (first-person) and GUI, including a message box to communicate with the experimenter, and an inventory.}
	\label{fig:client_view}
\end{figure}

In the corpus task, human participants played the role of a robot performing a situated reference resolution task. The study was run online using an interactive setup in which participants tele-operated a virtual PR2 robot.
The goal of the task was to organize a number of previously-unseen tools scattered around the spacecraft. A complete study run consisted of six \textit{trials}, each one corresponding to a single tool being located, then placed in the correct container. Participants used their keyboard to control the robot and pick up / place objects. They had a first-person view of the environment, along with an on-screen graphical user interface (GUI) that included a text box to communicate with a remote ``commander'' (an experimenter) through low-latency text chat (see Figure~\ref{fig:client_view}). The commander gave scripted instructions in each trial (to maximize experimental control) and responded to participant questions in a standardized manner.

\subsection{Environment Properties}
\label{sec:stimuli}
The environment consisted of an open room with various containers, interfaces, objects, and landmarks. The environment also contained six \textit{tool types}, with three \textit{instances} of each type. The tools were given fictitious names such that the names could not be used solely to identify the objects. The six tools were called: \textit{optimizer, calibrator, module, synthesizer, capacitor}, and \textit{emitter}. The tools had an additional prefix added to this label, resulting in object names like \textit{megaband module} and \textit{temporal emitter}. Only six instances were task-relevant, so the remaining twelve were distractors. 

In order to elicit a variety of question types, the tools were designed to be heterogeneous with respect to a number of features. Six features were assigned to the tool instances, including \textit{color (red, yellow, blue, etc.), shape (short, tall, narrow, wide), size (small, medium, large), texture (wood, coarse, metal, smooth), symbol (x, +, -, etc. on the object)}, and \textit{pattern (striped, spotted, no pattern)}. Each tool type varied along three feature dimensions, which resulted in a total of 20 combinations of three-feature sets \(\binom{6}{3}\).
The feature set of each of the six tool types was randomly selected from these combinations. For example, \textit{modules} varied by pattern, shape, and symbol whereas \textit{synthesizers} varied by color, size, and shape. 

In addition to the tools, the environment also contained a number of \textit{containers} (lockers, cabinets, and crates) and \textit{interfaces}, which were used to open the containers. The lockers, cabinets, and crates were locked and required learning a procedure to open (e.g., entering a passcode, arranging keycards, and pressing buttons). Various landmarks were placed throughout, including a wall with a transparent window, text on the walls (e.g., \texttt{X99}), electrical components on the walls, windows, and lights. These landmarks were put in place to elicit spatial referring expressions and other questions unique to situated environments. 

\subsection{Experimental Conditions}
The study used speaker initiative as a between-subjects factor, and instruction granularity as a within-subjects factor. The effect of these factors on question generation and performance is of particular importance since many human-robot interactions involve mixed-initiative dialogue \cite{mavridis2015review} and varying degrees of instruction granularity \cite{marge2020lets}. Our analysis was designed to investigate the effect that these factors have on the frequency and form of robot-initiated questions. 

Regardless of condition, all participants saw the same starting configuration of objects. The main difference was in the experimenter script and policy. In the Commander-Initiative (CI) condition, the commander provided each instruction (e.g., ``The sonic optimizer goes in the secondary cabinet, shelf A"), whereas in the Robot Initiative (RI) condition, the participant had to establish which tools and locations were relevant. 

Instruction granularity was manipulated by first preparing the high-level instructions and then converting these into low-level instructions by breaking them down into their constituent commands. For example, the high-level instruction, ``The tesla capacitor goes in the secondary cabinet, shelf B'' would become the following four low-level instructions: ``Move to the primary cabinet'' $\to$ ``Pick up the tesla capacitor from shelf A'' $\to$ ``Move to the secondary cabinet" $\to$ ``Place the tesla capacitor on shelf B''. In the CI condition, exactly three out of the six trials in each run were low-level instructions, and the other half were high-level instructions (the order was counterbalanced across participants).

\subsection{Predictions}
The following measures were used in the analysis: \textit{task performance} - based on the percentage of the six task-relevant objects placed correctly, \textit{task duration} - based on how long in seconds it took to complete the task, \textit{questions / total utterances} - which indicates the frequency of questions in the dialogue, and \textit{proportion of question types} - which indicates the relative frequency of each question type.

For our analysis, we had certain \textit{a priori} predictions. First, we expected that questions would be very common due to the need to clarify ambiguity. However, WHQs would be more common since the answers to them provide more information than a YNQ. We expected people to ask approximately three questions on average to resolve a referential ambiguity, corresponding with the three properties by which the entities varied. We also expected people to ask about object descriptions using salient properties like color and shape, though we did not know which properties would be salient. While object features are more salient if a single feature can be used to distinguish them (e.g., ``feature singletons'' from \cite{yantis2005visual}), we are not aware of studies finding greater salience of certain features in objects possessing heterogeneous features. 

We predicted that trials involving high-level instructions would take longer compared to low-level trials due to the more complex, and uncertain, instructions. This is because low-level instructions are very explicit and typically elicit only a single type of uncertainty. On the other hand, high-level instructions may have several unknown terms, and people need to decide on a method to clarify these terms. Regarding questions, the study was designed to elicit a broad range of question types. While the aggregate results will inform the kinds of questions people generally ask, we expected to see more questions overall with the high-level instructions due to the increased complexity. Finally, we predicted that speaker initiative would affect performance and efficiency in that people in the CI condition would perform better, finish the task faster, and ask fewer questions. This is because the task goals are explicitly laid out and structured when compared to the RI condition, in which participants do not have explicit goals.

\subsection{Empirical Results}
\label{sec:results}

The final dataset included dialogues from twenty-two participants, 10 in the CI condition and 12 in the RI condition. 11 of the participants were female and the rest were male. The average age was 37 $\pm$ 7 years and all participants were native English speakers. The mean number of questions to resolve an ambiguity was 1.72 $\pm$ 0.40, suggesting that people were very efficient at asking disambiguating questions. 

\subsubsection{Distribution of Questions}
\begin{figure}[t]
	\centering
	\includegraphics[width=\linewidth]{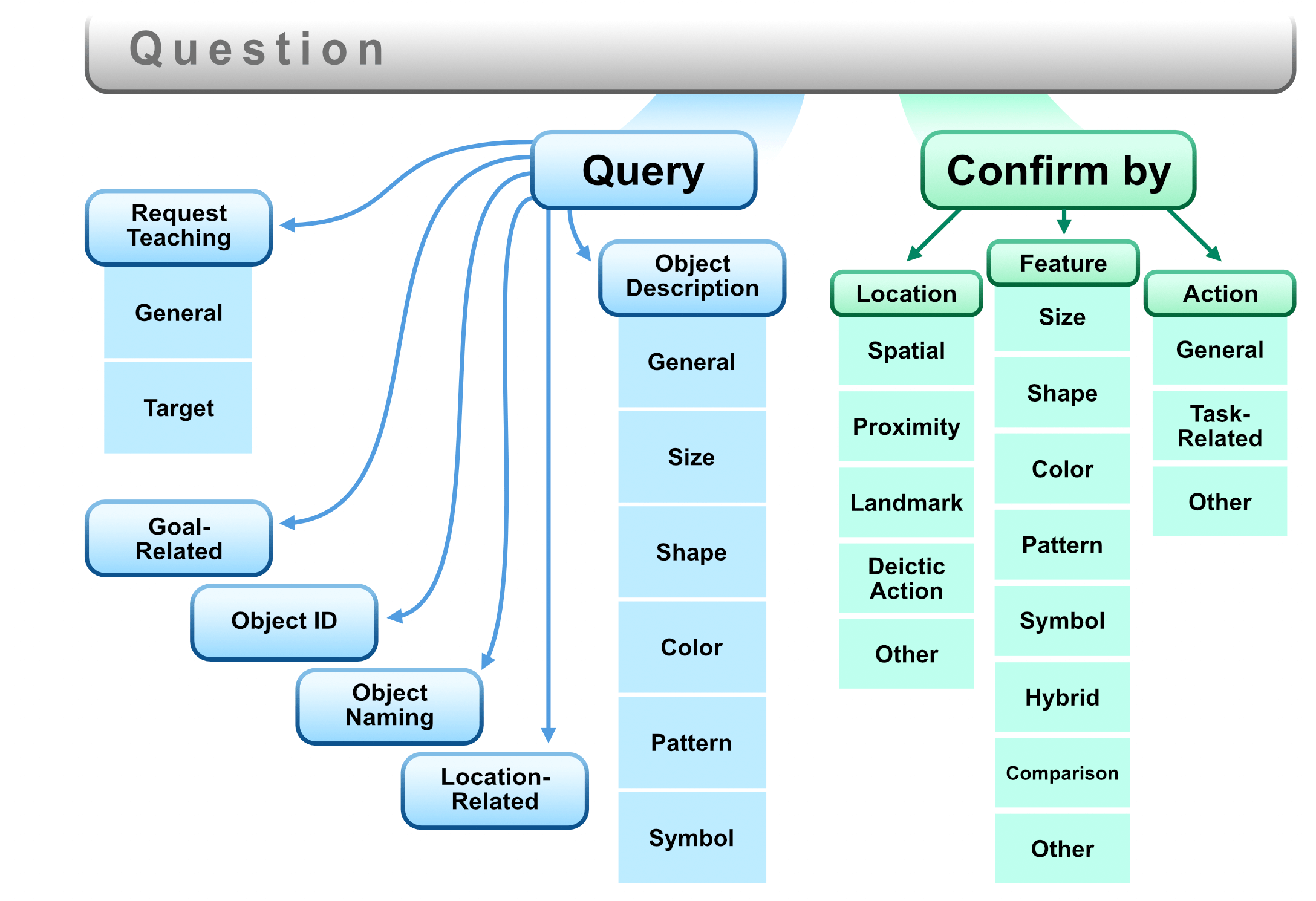}
	\caption{Question types used in the task. Queries generally corresponded to WHQs, whereas Confirm By's generally corresponded with YNQs.}
	\label{fig:question_types}
\end{figure}

We analyzed the transcribed dialogue data for the distribution of question types, which led to several interesting findings about how people ask questions to resolve referential ambiguity. As expected, we found that most utterances were questions, with 50\% in the form of WHQs, 15\% in the form of YNQs and 3\% as AQs (\textit{Alternative Questions}, which present a list of options); the rest were statements. This fit our prediction that WHQs would be the preferred question type. Though YNQs were less frequently used, feature-based confirmations were the most common subtype. In terms of feature preference, color was the most commonly asked-about feature, even though the objects were heterogeneous and only half of them varied by color. Among the location-based confirmations, spatial types were the most common followed by landmark types. Finally, action confirmations were mostly task-related (e.g., ``Does this go in the locker?''). Figure~\ref{fig:question_types} shows a taxonomy of question types observed in the data. These results will be used in Section~\ref{sec:model} for selecting questions and setting the utilities of the model.

\subsubsection{Instruction Granularity}
Within the CI condition\footnote{This analysis was run only on the CI condition because instruction granularity was not controlled in the RI condition.}, instruction granularity affected the trial duration and questions per utterance. Specifically, we found a difference in high- vs low-level instructions on trial duration and questions per utterance as indicated by paired-sample t-tests. High-level trials had a longer duration (\textit{M} = 1063.3 $\pm$ 274.2) than low-level (\textit{M} = 748.4 $\pm$ 228.3) trials, \textit{t}(9) = 3.04, \textit{p} $<$ .05. As we hypothesized, the less structured nature of high-level trials led to longer duration, suggesting that these instructions contained more uncertainty. In addition, high-level trials had a higher proportion of questions to utterances (\textit{M} = 0.74 $\pm$ 0.08) compared to low-level trials (\textit{M} = 0.58 $\pm$ 0.20), \textit{t}(9) = 2.26, \textit{p} $<$ .05 even though the average number of questions and utterances did not differ between Granularity conditions (\textit{p}s $>$ .05). This was an expected result given that low-level trials often had acknowledgments after each low-level command, and these were counted as \textit{Statements} in our scheme. There was no effect on performance or the average number of questions per trial (\textit{p}s $>$ .05).

\subsubsection{Speaker Initiative}
We ran a one-way ANOVA on dialogues from the 22 participants to examine the effect of speaker initiative. Contrary to our predictions, there was no effect of initiative on the number of questions, task performance, or task duration (\textit{p}s $>$ .05). We did however find significant differences in question types between the two Initiative conditions, specifically involving the proportion of \textit{Confirm Actions} / total questions [\textit{F}(1,20) = 4.69, \textit{p} $<$ .05] and \textit{Goal-Related} / total questions [\textit{F}(1,20) = 4.49, \textit{p} $<$ .05]. In both cases, the RI group displayed higher proportions of these question types, with \textit{Confirm Actions} making up 9.8\% of all questions and \textit{Goal-Related} making up 17.8\% of all questions (compared to 1.7\% and 8.5\% in the CI condition, respectively). As predicted, the lack of guidance in the RI condition led to the increased prevalence of these question types. 

\section{Decision-Theoretic Model}
\label{sec:model}

In this section we introduce a computational model of dialogue-based situated reference resolution that is informed by the empirical findings described in the previous section. The model uses a decision network \cite{owen1978use} to represent the agent's knowledge about relevant properties of the entities in the world (see Section~\ref{sec:decision_net}). This knowledge can be deterministic (e.g., $\mathit{knows}(x)$) or probabilistic (e.g., $\mathit{knows}(x)_{p = .8}$). Using this approach, the agent selects and asks a single question about some property that maximizes its expected utility over all possible questions. It then interprets the response and continues asking questions in this way until it has resolved the referent. The list of question types was based on a subset of the ones in Figure~\ref{fig:question_types}, and the properties and initial utility table also came from the task environment.

While the structure of the decision network implementation reflects the particular properties of our task domain (e.g., shape, size, color, etc.), these are general object properties that can be used to disambiguate entities in a variety of domains. The approach, however, does not depend on these particular properties, and works for any arbitrary number and type of properties. The question types are also domain-independent, and are based on the properties. Moreover, we can construct the decision network in real time so that it can be used in a variety of domains without prior training (see Section~\ref{sec:dynamic_model}).

\subsection{Decision Network Specification}
\label{sec:decision_net}

Decision networks are probabilistic graphical models related to Bayesian networks that encode conditional relationships between variables in a directed acyclic graph. They are effective at integrating multiple sources of uncertainty (e.g., vision, language, knowledge) and supporting probabilistic decision-making. Decision networks contain three types of nodes: \textit{chance nodes} which are random variables, \textit{decision nodes} which reflect actions that an agent can perform, and \textit{utility nodes} which represent the utility function over those actions. The goal of the network is to find the action with the maximum expected utility (MEU) using Equation~\ref{eq:MEU}. To compute the expected utility of each action $a$ (i.e., question), given evidence $e$, and all possible values of the chance nodes, s, the network multiplies the posterior probabilities, $P(s|e)$, by the utility weights for that action, $U(a,s)$. The posterior probabilities start out uniformly distributed, and are updated as the agent perceives and learns about additional object properties. The utility weights are set using an entropy-driven approach (see ~\ref{sec:utility} below). 

\begin{equation}
\label{eq:MEU}
\mathit{MEU}(e) = \argmax_a{\sum_s{P(s|e)U(a,s)}}
\end{equation}

\begin{figure}[t!]
	\centering
	\includegraphics[width=8cm]{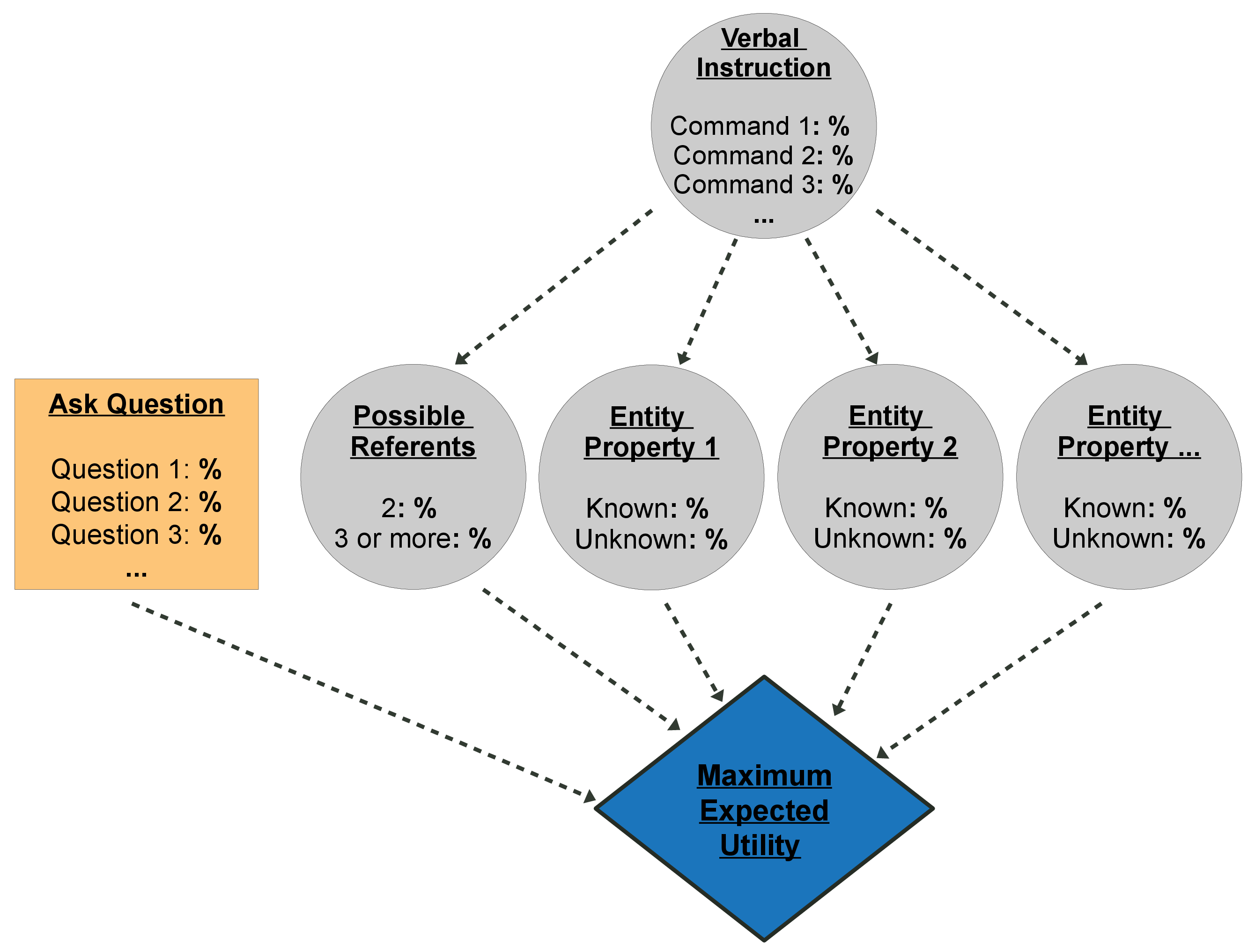}
	\caption{Decision Network to generate questions for situated reference resolution. The grey circles are chance nodes (random variables), the orange square is a decision node, and the blue diamond is a utility node. See text for more details. }
	\label{fig:decision_network}
\end{figure}

The general structure of the decision network is depicted in Figure~\ref{fig:decision_network}. The network contains the following node types: 

\subsubsection{Chance Nodes}
These are random variables that represent (1) the possible verbal instructions, (2) the number of possible referents that correspond to the input command, and (3) the agent's knowledge about the various properties that entities can possess in the environment. A separate node is used for each entity property, and represents a probability function of the agent's knowledge of that property given the input commands. The prior probabilities for the \textit{verbal instructions} chance node are uniformly distributed over the vocabulary of instructions in the task domain, and are updated when an instruction is received by the system. The prior probabilities for the \textit{referents} and \textit{properties} chance nodes are also uniform, but these values are updated based on the agent's knowledge acquired during the task.  

\subsubsection{Decision Node} 
A decision node is a list of questions that the agent can ask. We selected the most frequently used twelve question types from the corpus, though the node can support an arbitrary number of questions. Nine WHQs were used, including all of the Object-Description queries, and three YNQs were selected, from those used at $>$1\% frequency in the data (\textit{Confirm:Color, Confirm:Spatial, Confirm:Landmark}). 


\subsubsection{Utility Node} 
\label{sec:utility}
A utility node is a conditional probability table that represents the utility of asking a question given the agent's knowledge of the various properties and the number of ambiguous referents that correspond to the input. We explore two ways of setting these utilities, though others may also be used. The first is a \textit{data-driven} method that involves using the corpus data to set the utilities. Specifically, we used the relative frequency (normalized) of each question type. The procedure sets the data-derived utility value for all \textit{unknown} properties that correspond with that question, whereas the \textit{known} properties are set to 0. For example, the utility of \textit{Query:Shape} when knowledge of \textit{Shape} is unknown would be the data-driven value (e.g., 20), and otherwise it would be 0. This approach creates a natural preference ranking over the questions. 

The second, and more general, method for setting utilities is an \textit{entropy-driven} approach based on computing the Shannon entropy for each property. For WH-questions about a property $\phi$ (e.g., ``What color is it?''), with values $\phi_1,...,\phi_n$ (e.g, red, green, blue, etc.), this utility is: 
\begin{equation}
\label{eq:WHQentropy}
U_{\mathit{WH}}(\phi) = -\sum_{i=1}^{n} P(\phi_i)\cdot log P(\phi_i) 
\end{equation}

\noindent For YN-questions about a property $\phi$ (e.g., ``Is it red?''), the system will learn either that the object does or does not have a particular property value. Therefore, we computed the expected entropy for YNQs from individual entropies associated with each possible value of the property: 
\begin{multline}
\label{eq:YNQentropy}
U_{\mathit{YN}}(\phi) = \sum_{i=1}^{n} P(\phi_i)\cdot(-P(\phi_i)\cdot log P(\phi_i)\\
-P(\neg\phi_i)\cdot log P(\neg\phi_i)) 
\end{multline}

\subsection{Dynamic Model Generation}
\label{sec:dynamic_model}
To support generalizability, we developed a procedure to dynamically construct the decision network from a given world state in real-time. First, the world state is processed based on the agent's perception. This includes entities, their properties and values, and a mapping of properties to questions. Next, we compute the \textit{minimum disambiguating set} of properties of the world to determine the number of questions and structure of the decision network (see Algorithm~\ref{alg-minimum-props}). The entropy of all entity properties is then computed, and a utility table is constructed based on the entropy values in Equations~\ref{eq:WHQentropy} and~\ref{eq:YNQentropy}. Finally, the decision network is instantiated with the above parameters. 

\begin{algorithm}
\caption{Algorithm to compute the minimal disambiguating set of properties.}\label{alg-minimum-props}
\begin{algorithmic}[1]
\Procedure{ComputeMinSet}{$entities$}
\State $C \gets list$
\State $H \gets list$
\State $propertyList \gets list$
\State $entityList \gets combinations(entities)$
\\
\ForAll {$e1, e2$ in $entityList$}
    \ForAll {$p$ in $e1.properties$}
        \If{$e1.p$ $\neq$ $e2.p$}
            \State $propertyList$.append($p$)
        \EndIf
    \EndFor
    \State $C$.append(CNF($propertyList$))
\EndFor
\\
\State $H \gets SAT$.solve($C$)
\State \textbf{return} $H$
\EndProcedure
\end{algorithmic}
\end{algorithm}

To compute the minimum disambiguating set of properties\footnote{For example, if two objects have the same shape and size, but differ in color, then the minimum set is \textit{\{color\}.}}, we formulate the problem as an instance of the well known Hitting Set (Set Cover) problem and use a SAT solver to find a solution (or approximation in larger domains). The process is detailed in the algorithm in Algorithm~\ref{alg-minimum-props}. \textit{C} is a collection of subsets (from a domain) that represents the properties by which all entities differ from one another. \textit{C} is computed in lines 7-12 by generating all pairs of entities and comparing each of their properties. The resulting list of disambiguating properties is converted to a Boolean formula in conjunctive normal form (CNF) and added to \textit{C} (line 13). Then we compute the smallest subset \textit{H} by running a SAT solver on \textit{C} and returning the minimum disambiguating set. This process enables an agent to disambiguate every entity in the environment based on their properties. 

\section{Model Evaluation}
\label{sec:evaluation}

\begin{table*}[t]
\centering
\caption{Evaluation results for reference resolution efficiency of unknown entities (mean questions per instruction). Lower values indicate greater efficiency. The best results in each environment are bolded.}
\label{tab:eval_results}
\begin{tabular}{lllllllll} \toprule
\textbf{}              & \multicolumn{2}{c}{\textbf{Human (Corpus)}} & \multicolumn{2}{c}{\textbf{Baseline}} & \multicolumn{2}{c}{\textbf{Model (Data)}} & \multicolumn{2}{c}{\textbf{Model (Entropy)}} \\ \midrule  
\textbf{Environment}   & \textbf{M}      & \textbf{SD}      & \textbf{M}        & \textbf{SD}       & \textbf{M}          & \textbf{SD}         & \textbf{M}           & \textbf{SD}           \\ \midrule
Spacecraft Domain           & \textbf{1.72}            & 0.4              & 2.75              & 0.33              & \textbf{1.75}                & 0                   & \textbf{1.75}                 & 0                     \\
Random (low variance)  & -               & -                & 5.09              & 0.71              & 1.91                & 0.14                & \textbf{1.80}                 & 0.09                  \\
Random (high variance) & -               & -                & 2.66              & 0.37              & 1.96                & 0.19                & \textbf{1.85}                 & 0.13          \\        \bottomrule
\end{tabular} 
\end{table*}

To evaluate our proposed approach, we created a framework for simulating a situated dialogue task. In particular, we simulated the process of reference resolution that would arise when a situated agent would be asked to carry out a command involving an unknown entity, such as in the corpus task. \textbf{The principle aim of the simulation framework was to evaluate the efficiency of our proposed model, namely the number of questions required to correctly identify and resolve a reference to an unknown entity.}\footnote{Note that this is different from a typical reference resolution evaluation (e.g., \cite{kennington2017simple}) in that the focus here is on dialogue efficiency, not resolution accuracy.} 

The task world consisted of a set of entities, each with various properties. Each entity was unique, such that no two objects possessed the same set of property values. Task world configurations can be manually specified, such as in the case of the spacecraft domain, or randomly generated. Random generation of task worlds allows us to evaluate how well our proposed system would generalize to a wider array of possible situations. In particular, to evaluate the ability of our question-asking system to ask well-targeted questions, we created two types of randomly generated worlds: (1) {\em low property variance} where entity property values varied only across a small (\textit{N=3}) set of properties; and (2) {\em high property variance} where entity property values varied across a large (\textit{N=7}) set of properties. 

\subsection{Simulated Dialogue Agent}

We implemented a simulated dialogue agent that was tasked with resolving the identity of an unknown entity through dialogue. The agent could ask WHQs and YNQs regarding particular object properties to an oracle agent, which would respond based on the ground truth from the task world configuration. The simulated dialogue agent would then reduce the set of possible candidate objects based on this information. The evaluation trial would end once the dialogue agent successfully reduced the set of candidate objects to one. The number of questions required during each trial was recorded.

Two models were implemented and compared in the evaluation: (1) the model from Section~\ref{sec:model} (with two utility assignments) and (2) a slot-filling baseline. The baseline asks questions corresponding to unknown properties of the entities. Once a property is learned (e.g., \textit{color:red}), it no longer asks questions about that property. To choose a question, the baseline randomly selects from the same set of questions available to the model. This stochasticity makes the baseline approach effective in the random worlds since the relevant properties are not known in advance. 

We were interested in comparing how well the model performs relative to this baseline and to the humans in the corpus. The human comparison informs the feasibility of this approach as a cognitive model of human question generation under uncertainty. We predicted:\\
\noindent \textit{(\textbf{H1}): The model will outperform the baseline in the spacecraft domain}. \\
This is because the model's utilities are set to prioritize salient object properties that humans queried to succeed in the task. We also expected the model (but not the baseline) to achieve similar performance to humans in the corpus.\\
\noindent \textit{(\textbf{H2}): The model will outperform the baseline in the random domains, both for high and low property variance}. \\
However, this advantage would only hold when the decision network utilities were entropy-driven rather than data-driven, since the data-driven utilities reflect a static configuration of only one environment. \\
\noindent \textit{(\textbf{H3}): Systems will perform better in domains with high property variance}.\\
To understand the effect of property variance, we performed within-model comparisons in the random environments. Since higher variance leads to more opportunities to ask good questions, there is a higher chance that a question will target a disambiguating property.

\subsection{Evaluation Results}

Three systems were evaluated: baseline, entropy-driven model, and data-driven model. 20 trials (i.e., instructions) were provided to each system, each corresponding to an entity, e.g., ``Pick up the temporal emitter''\footnote{These trials correspond to the Commander Initiative, low-level Instruction Granularity condition in the corpus.}. We ran 100 iterations of these trials in each comparison, resulting in 2000 entities to resolve for each system. For the runs in the spacecraft domain, the environment layout from the corpus was used. For the random domain runs, a new world state (and decision network) was randomly generated with each iteration according to the property variance constraints. Independent-samples t-tests were used to evaluate \textit{H1} and \textit{H2}, and paired-samples t-tests were used to evaluate \textit{H3}. The dependent variable in all analyses was \textit{mean questions per instruction} (averaged across all 100 iterations), which reflects question generation efficiency. This was used as the primary measure because resolution accuracy was always 100\%, as the oracle agent had ground truth knowledge, and the correct referent could always be identified with enough questions.

Our results supported \textit{H1} in that the model showed greater efficiency in question generation than the baseline system in the spacecraft domain (see Table~\ref{tab:eval_results}). Specifically, the model with data-driven utilities asked fewer questions than the baseline to resolve all entities, \textit{t}(198) = 30.52, \textit{p}$<$.001. As predicted, there was no difference in performance of the model compared to the human data (\textit{t}(108) = 0.70, \textit{p}=.48), however, the humans outperformed the baseline system, \textit{t}(108) = 9.25, \textit{p}$<$.001.

We found partial support for \textit{H2}. As predicted, the model with entropy-driven utilities outperformed the baseline in domains with both low (\textit{t}(198) = 46.03, \textit{p}$<$.001) and high (\textit{t}(198) = 20.40, \textit{p}$<$.001) property variance. However, contrary to our prediction, we found that this advantage held even when the model used data-driven utilities across both property variance conditions (\textit{ps}$<$.001). This result was likely driven by the minimal disambiguating set calculation, as it served to reduce the space of questions sufficient to maintain efficiency. 

Finally, we found partial support for \textit{H3} in that the Baseline system performed significantly better with high property variance, but the Model performed significantly worse (\textit{ps}$<$.05). The results for the Baseline system are intuitive in that domains with a higher degree of property variance are simpler for reference resolution since questions about most properties will be informative. However, though the Model was significantly better in the high property variance condition, the means are much closer than for the Baseline system (low: 1.91 vs high: 1.96). The difference is likely explained by the use of Algorithm~\ref{alg-minimum-props}, which reduced the number of properties, and hence mitigated the drawbacks of low property variance.


\section{Discussion}
\label{sec:discussion}

\subsection{Contributions}
\subsubsection{Scientific Contributions}
Our empirical results extend prior work on clarification in human-human and human-agent dialogue in the presence of multi-modal uncertainty. First, they highlight the distribution of question types that people use to resolve referential ambiguity, suggesting that WHQs, and particularly feature queries, are very common. By contrast, the results from \cite{cakmak2012designing} on a human learning task (using different annotation categories) found that 81\% of questions were YNQs, 13\% were WHQs, and 4\% were AQs. This reflects the different domains and familiarity with the task. Feature-based questions were extremely common in their corpus (82\% of all questions), whereas we observed about 20\% of questions this category. Like in their analysis, many of the questions we observed were grounded in properties of the entities and environment. This extends prior work on visual salience \cite{misu2015visual,yantis2005visual}, informing which properties are preferred in feature-heterogeneous entities. 
Second, the empirical results shed light on the impact of dialogue-level factors such as speaker initiative and instruction granularity on question generation. Interestingly, none of these factors impacted task performance. In terms of task duration, however, trials with high-level instructions were found to take longer on average to resolve than low-level instructions. This was likely due to the fact that high-level instructions had more uncertainty packed into them (i.e., identifying the object and the location), thus requiring more processing to resolve. As a result, high-level trials also elicited a greater proportion of questions than low-level trials, which likely reflected the extra uncertainty. In terms of speaker initiative, we found several differences in question types between the CI and RI conditions. In particular, the RI group used more \textit{Confirm Actions} and \textit{Goal-Related} queries than the CI group. These question types reflect the additional ``burden" of taking initiative, and highlight required capabilities of robots in this domain.

\subsubsection{Technical Contributions}
The computational model of situated reference resolution employs an empirically-inspired decision-theoretic approach to resolving referential ambiguity through dialogue. The approach is well suited to handling uncertainty across multiple modalities, including vision, speech, knowledge, and others. In an evaluation with a simulated dialogue agent, the model outperformed a baseline system in terms of question efficiency, and showed no difference from human-level performance. The similarity to the human data suggests that the model may be capturing some of the salience-based mechanisms which humans utilize to prioritize their questions. However, one important difference is that the model used much fewer YNQs than humans, due to the reduced entropy of such questions compared to WHQs. While more work is needed, this model serves as a step toward a computational cognitive model of clarification under uncertainty. 

The advantage of the model over the baseline system stems from several areas. First, the utilities enable it to prioritize more informative questions. By effectively ranking the highest entropy properties in the environment, the decision network encodes the most relevant properties to ask about. These properties include those unique to situated interaction, such as spatial orientation and landmark proximity. Second, the model is more efficient because it constrains its space of questions based on the minimum disambiguating set of properties. This, combined with the entropy-driven utilities, enables the model to scale to larger domains and to ask the most informative questions each time. Finally, dynamic construction of the network enables generalizability to open-world domains with no prior training necessary. This is a key advantage over existing approaches that require large amounts of training data and are limited to closed worlds. 

Overall, this model can integrate well with a variety of systems, especially those that ground perceptual input into symbolic representations, such as \cite{bullard2018towards} and \cite{williams2019dempster}. These systems can benefit from the model's focus on property-related questions, with the symbol groundings serving as input to the model. The model can also be used to enhance POMDP-based dialogue systems (e.g., \cite{lison2013model}), which excel at dialogue state tracking, but struggle with scalability and open-world domains. 

\subsection{Limitations and Future Work}

One limitation of the analysis and model is the possibility that the results will not generalize to physical domains or other kinds of tasks. Though the corpus task involved a virtual robot, the interaction was situated in an environment in order to elicit uncertainty in locations, spatial reference, and procedural knowledge - all properties highly relevant for physical robots as well. We do not see any reason why the analysis and the model would not extend to physical robots, though of course such robots will need to contend with additional factors such as noisier sensor readings and potential task failure. Though the model was designed to be generalizable, future studies with physical agents in other task domains will be needed to replicate our findings and explore these additional factors.

Directions for future work are two-fold. First, we intend to further analyze the empirical dataset to better understand the questions that people asked and how these influenced task performance and efficiency. This may yield new findings of factors that influence question generation, leading to refinements of the model. The second direction for future work is to improve the computational model. We will investigate stochastic domains in which the agent has multiple sources of uncertainty and limited knowledge of the entities and their properties. The decision network is naturally designed to handle uncertainty so it would be informative to evaluate it in these more realistic domains. Finally, we will integrate this model within the language pipeline of a robot architecture and test performance in real, physical environments where world knowledge comes from noisy sensor data.

\section{Conclusion}
\label{sec:conclusion}

We presented a decision-theoretic approach to situated reference resolution for dialogue agents that was grounded in empirical findings from a corpus of clarification dialogues. The analysis revealed novel findings about the effect that dialogue-level factors have on question generation under multi-modal uncertainty. The results also informed the distribution of question types that humans use when resolving novel entities, and led to a computational model for automated question generation. The model uses a decision network and an entropy-based utility assignment method to generate efficient questions for reducing referential ambiguity. Moreover, it can be generated in real-time, making it generalizable to open-world domains. This approach can be integrated into a variety of dialogue systems to perform efficient reference disambiguation in open worlds, and is particularly well suited to multi-modal domains. Overall, these results offer insight into the ways that agents can ask questions to facilitate learning, and serve as a step toward improving automated approaches for situated reference resolution.

\begin{acks}
This research was sponsored by the Basic Research
Office of the U.S. Department of Defense with
a Laboratory University Collaboration Initiative
Fellowship awarded to MM, and an Air Force Office of Scientific Research grant to MS under award FA9550-18-1-0465. 
\end{acks}

\bibliographystyle{ACM-Reference-Format}
\bibliography{refs}


\end{document}